\documentclass[a4paper,fleqn]{cas-sc}
\usepackage[numbers]{natbib}
\usepackage{subfig}
\usepackage{graphicx} 
\usepackage{diagbox}
\usepackage{multirow}
\usepackage{booktabs}
\usepackage{amsmath}
\usepackage{amssymb}
\usepackage{algorithm}
\usepackage{soul,color}
\usepackage{algpseudocode}
\usepackage{hyperref}
\usepackage{pifont}
\def\tsc#1{\csdef{#1}{\textsc{\lowercase{#1}}\xspace}}
\tsc{WGM}
\tsc{QE}
\tsc{EP}
\tsc{PMS}
\tsc{BEC}
\tsc{DE}

\begin{document}
\let\WriteBookmarks\relax
\def\floatpagepagefraction{1}
\def\textpagefraction{.001}

\shorttitle{Tri-CRLAD Based on Causal Inference for Sensor Signals}

\shortauthors{X.Chen R.Xiao et~al.}

\title [mode = title]{Semi-supervised Anomaly Detection via Adaptive Reinforcement Learning-Enabled Method with Causal Inference for Sensor Signals}




%
\author[1,2]{Xiangwei Chen}
	\ead{aoudsung@gmail.com}
	
	\author[1,2,3]{Ruliang Xiao}
	\ead{xiaoruliang@fjnu.edu.cn}

    \author[1]{Zhixia Zeng}
	\ead{QBX20220156@yjs.fjnu.edu.cn}
  
    \author[1]{Zhipeng Qiu}
      \ead{qbx20230179@yjs.fjnu.edu.cn}
      
    \author[1]{Shi Zhang}
      \ead{shi@fjnu.edu.cn}

     \author[1]{Xin Du}
      \ead{xindu79@126.com}
      
\address[1]{College of Computer and Cyber Security, Fujian Normal University, Fuzhou Fujian 350117, China}
\address[2]{Fujian Provincial Digit Fujian Internet- of Things Laboratory of Environmental Monitoring, Fuzhou Fujian 350117, China}
\address[3]{Fujian Provincial Key Laboratory of Network Security and Cryptology, Fuzhou Fujian 350117, China}




\cortext[cor1]{Corresponding author:xiaoruliang@fjnu.edu.cn}



\begin{abstract}
Semi-supervised anomaly detection for sensor signals is critical in ensuring system reliability in smart manufacturing. However, existing methods rely heavily on data correlation, neglecting causality and leading to potential misinterpretations due to confounding factors. Moreover, while current reinforcement learning-based methods can effectively identify known and unknown anomalies with limited labeled samples, these methods still face several challenges, such as under-utilization of priori knowledge, lack of model flexibility, and deficient reward feedback during environmental interactions. To address the above problems, this paper innovatively constructs a counterfactual causal reinforcement learning model, termed Triple-Assisted Causal Reinforcement Learning Anomaly Detector (Tri-CRLAD). The model leverages causal inference to extract the intrinsic causal feature in data, enhancing the agent's utilization of prior knowledge and improving its generalization capability. In addition, Tri-CRLAD features a triple decision support mechanism, including a sampling strategy based on historical similarity, an adaptive threshold smoothing adjustment strategy, and an adaptive decision reward mechanism. These mechanisms further enhance the flexibility and generalization ability of the model, enabling it to effectively respond to various complex and dynamically changing environments. Experimental results across seven diverse sensor signal datasets demonstrate that Tri-CRLAD outperforms nine state-of-the-art baseline methods. Notably, Tri-CRLAD achieves up to a 23\% improvement in anomaly detection stability with minimal known anomaly samples, highlighting its potential in semi-supervised anomaly detection scenarios. Our code is available at \url{https://github.com/Aoudsung/Tri-CRLAD/}.
\end{abstract}




\maketitle
\section{Introduction}
In the era of smart manufacturing, growing sensing and computing capabilities have facilitated the development and monitoring of sensor signals in engineering and the application of anomaly detection. Anomaly detection, often referred to as the process of identifying deviations from the norm, plays a pivotal role in a multitude of sensor-based monitoring processes, including but not limited to healthcare surveillance, satellite monitoring, cybersecurity, and wind turbine monitoring \cite{pang2021deep,10496248_pami, DELIMAMUNGUBA2024108363-EA}. This field is highly sought after due to its inherent capability to enhance the reliability of the data collected, thereby contributing to the overall reliability and robustness of systems \cite{SARTOR2024107671-EA2, ZHAO2023106955_EA3, AKAGIC2024107368_EA4}.

However, since anomaly data is less than normal data, and the types of anomalies in real environments are complex and variable, which makes it is difficult to label them completely \cite{9492295} in real scenarios. Semi-supervised anomaly detection has received much attention from researchers because it requires less labeled data and can be combined with many unlabeled data streams to identify anomalies effectively. Hence, in intricate industry environments, the method's design should be strong enough to develop an efficient detection model with a restricted number of anomaly data and a large number of normal samples. Moreover, it should be capable of generalizing and detecting unknown anomalies.

Deep Reinforcement Learning (DRL) can effectively leverage its characteristics to meet the needs of semi-supervised anomaly detection methods \cite{chen2023deep, pang2021toward, watts2022dynamic}. The distinctive feature of DRL is its optimization of the decision-making process through balancing exploration and exploitation. Where exploration represents the ability of the agent to improve and refine its execution policy, and exploitation represents the ability of the agent to utilize the knowledge it has already learned to make decisions. Due to this characteristic, DRL has been widely used in several domains, such as autonomous driving, Cyber-Physical Energy System Operations and Maintenance, and robot control \cite{kiran2021deep,pan2022asynchronous,avramelou2024deep, HAO2023109231_rs}. For the semi-supervised anomaly detection problem, the DRL-based anomaly detection method learns efficiently with a small number of labeled anomaly samples and continuously explores and adapts to new anomaly patterns in real-time data streams.
\begin{figure}[ht]
    \centering
    \scalebox{1.0}{
        \begin{minipage}{\linewidth}
            \centering
            \subfloat[]{
                \includegraphics[width=0.48\linewidth, height=0.25\linewidth, keepaspectratio]{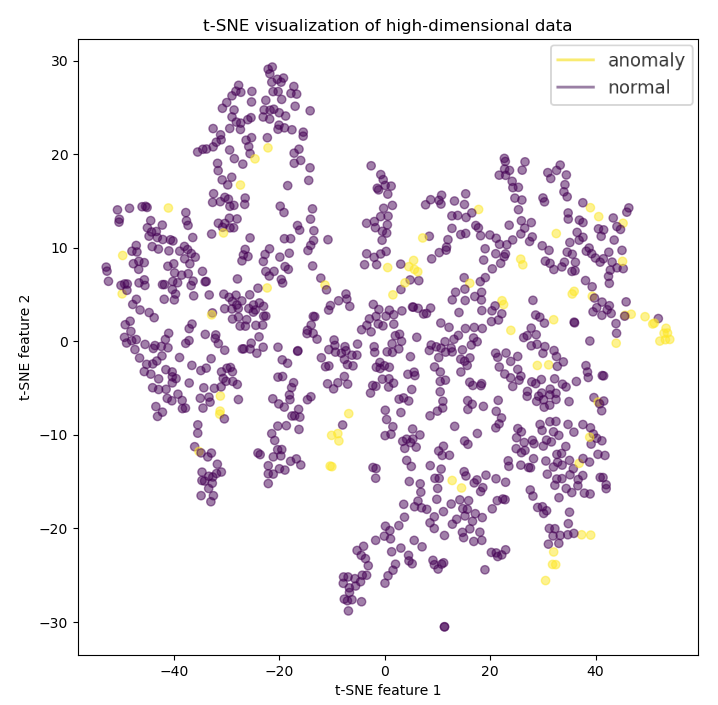}
                \label{fig1a}
            }
            \subfloat[]{
                \includegraphics[width=0.48\linewidth, height=0.25\linewidth, keepaspectratio]{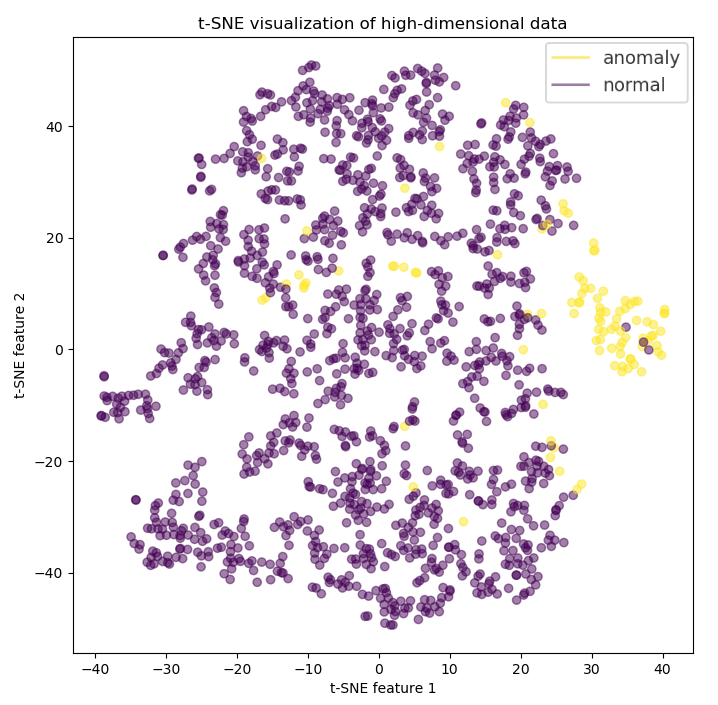}
                \label{fig1b}
            }
            \vspace{0.2em} 
            \subfloat[]{
                \includegraphics[width=0.48\linewidth, height=0.25\linewidth, keepaspectratio]{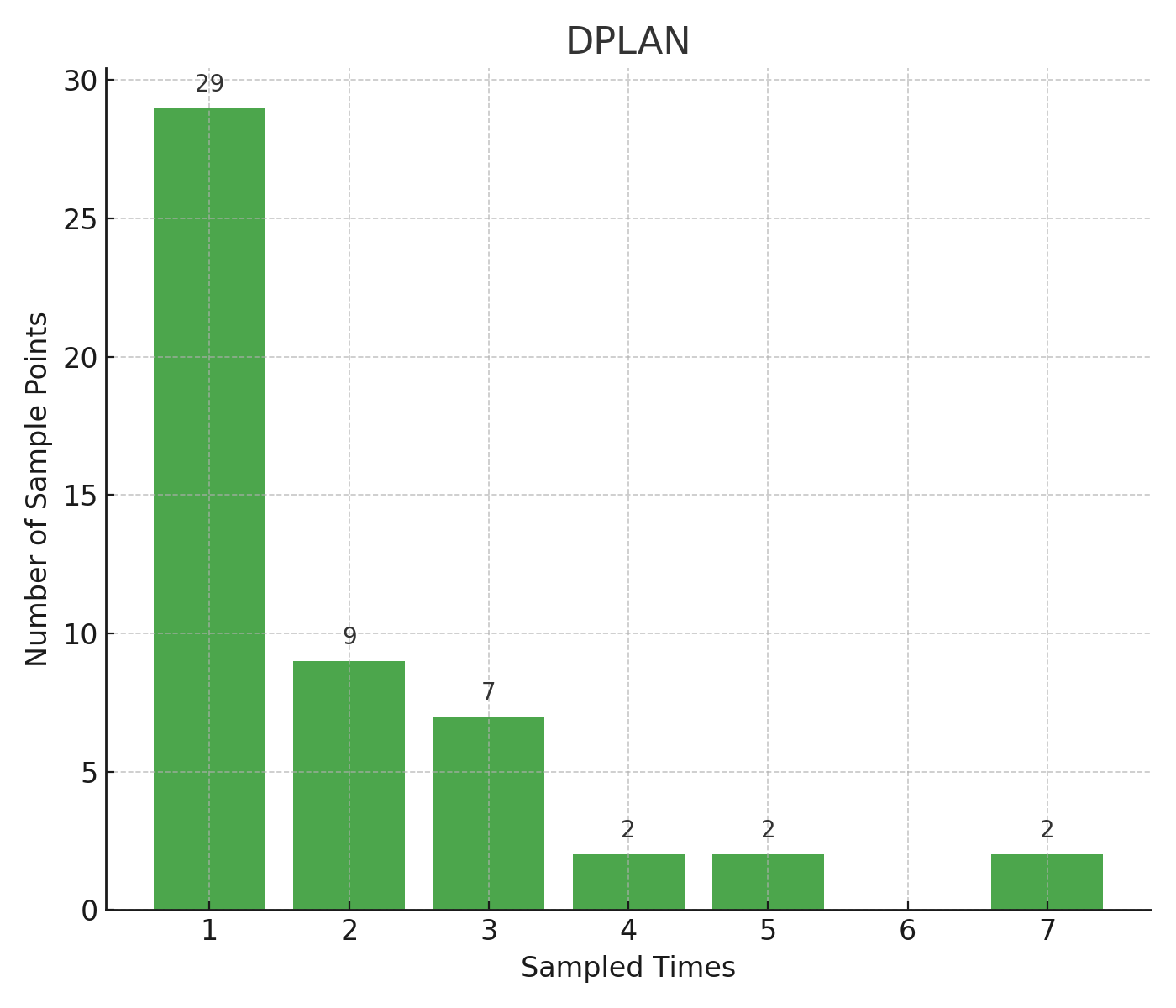}
                \label{fig1c}
            }
            \subfloat[]{
                \includegraphics[width=0.48\linewidth, height=0.25\linewidth, keepaspectratio]{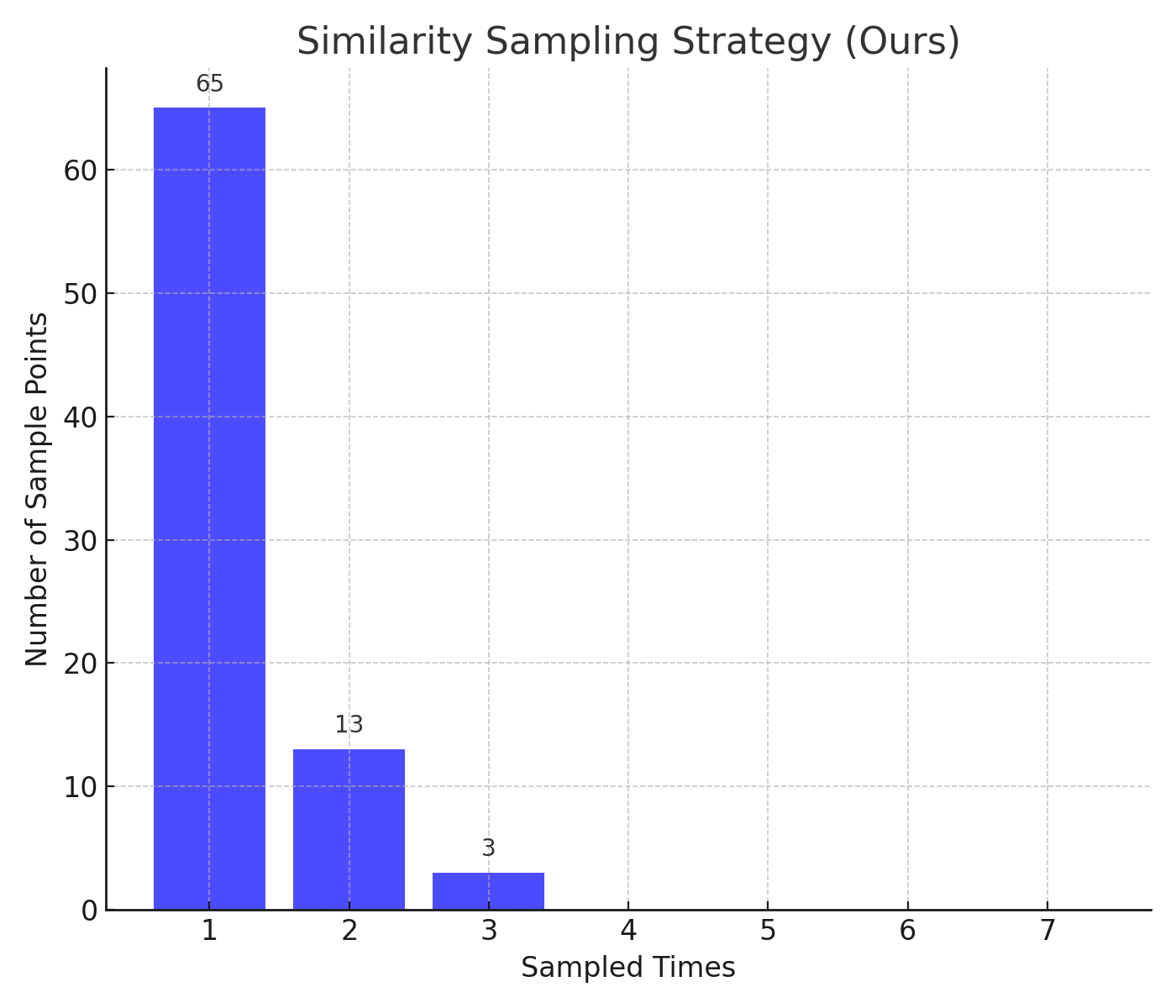}
                \label{fig1d}
            }
            \caption{Visualization of the effects of causal inference and sampling strategies. Subfigures \ref{fig1a} and \ref{fig1b} present the original versus causally inferred data features using t-SNE. Subfigures \ref{fig1c} and \ref{fig1d} depict the agent's frequency of visits to the same point over identical step counts, employing distance-based and historical similarity-based sampling strategies. The x-axis represents visit frequency, and the y-axis shows the count of points with equal visits. A higher y value at a lower x value suggests a superior sampling strategy. Subfigures \ref{fig1a} and \ref{fig1b} highlight the efficacy of causally processed features in identifying anomaly samples. Meanwhile, subfigures \ref{fig1c} and \ref{fig1d} demonstrate the proposed sampling strategy's effectiveness in facilitating a broader exploration of data points within a fixed step count, reducing the redundancy of revisiting identical points.}
        \end{minipage}
    }
    \label{fig.1}
\end{figure}
Given the high degree of fit between deep reinforcement learning's capabilities and the demands of semi-supervised anomaly detection, researchers have increasingly focused on the practical applications of DRL in anomaly detection \cite{chen2023deep, pang2021toward, ijcai2023p19}. However, these past approaches still suffer from the following problems:

 \textbf{Problem 1}: \textbf{Underutilization of prior knowledge}. Much of the previous work has focused on learning by utilizing correlations in the data rather than by utilizing the intrinsic causal relationships between sensor signal data. However, as shown in Figure \ref{fig1a}, the raw data features may not represent the difference between abnormal and normal data well, implying that abnormal data are not well identified by correlations alone. This makes the model susceptible to the adverse effects of confounding information (e.g., noise dimensions) present in the data, which reduces the model's reliability. On the other hand, the limitation imposed by the inadequate sampling method also restricts the utilization of prior knowledge. As shown in Figure \ref{fig1c} and \ref{fig1d}, when the x value is lower, a higher y value indicates a better sampling strategy. Figure \ref{fig1c} illustrates outcomes from the DAPLN method's distance-based sampling strategy, which was applied to 100 samples \cite{pang2021toward}. This approach risks model overfitting by repeatedly selecting the same data points, potentially diminishing the model's performance. In contrast, as shown in Figure \ref{fig1d}, our proposed sampling strategy then explores the entire dataset well.
 
\textbf{Problem 2}: \textbf{Fixed thresholds for anomaly determination limit the model's generalization ability and increase the complexity of parameter tuning}, such as the DADS \cite{chen2023deep}.

\textbf{Problem 3}: \textbf{Current reward mechanisms in environments with anomalous interactions are restrictive and fail to furnish adequate feedback.} These methods \cite{pang2021toward,ijcai2023p19} rely on a rudimentary value feedback mechanism, such as (-1, 0, 1). This mechanism is inflexible and does not allow for dynamic adjustment of rewards in response to the prevailing sensor signal context. When the scenario involves a scarcity of labeled data instead of an abundance, the probability of the intelligence encountering an anomaly data point diminishes. Consequently, if the reward system is confined to a narrow spectrum of situations, it compels the agent to explore a greater variety of behaviors, extending the exploration phase and impairing the model's learning efficiency.

In order to address the above problems, this paper introduces the Tri-CRLAD model. Tri-CRLAD developed a causal inference model comprising two stages, intra-point and inter-point modeling, accompanied by a triple decision support mechanism rooted in historical information. Intra-point modeling focuses on analyzing dimensional correlations within the sensor signal data. In contrast, inter-point modeling utilizes counterfactual causal inference to extract the intrinsic causal feature, thereby augmenting the DRL agent's effectiveness in applying the prior knowledge. In addition, the triple decision support mechanism skillfully integrates environmental changes into the decision-making process, enabling the agent to better adapt to changes in the ever-changing detection environments and thus effectively deal with the problem of semi-supervised anomaly detection for tabular sensor signal data. In summary, this paper makes the following contributions:
\begin{enumerate}[1)]
    \item Innovatively integrates causal reinforcement learning into semi-supervised anomaly detection, developing the Tri-CRLAD that employs counterfactual causal inference to identify intrinsic causal features in the data. This procession enhances the DRL agent's use of prior knowledge and its generalization capability.
    \item Introduces a comprehensive triple decision support mechanism based on historical information. This mechanism includes a sampling strategy grounded in historical similarity, an adaptive threshold smoothing adjustment strategy, and an adaptive decision reward mechanism. These innovations collectively tackle the identified challenges and significantly improve the Tri-CRLAD's training efficiency and generalization ability.
    \item Extensive experiments demonstrate that the Tri-CRLAD achieves performance that meets or exceeds the current state-of-the-art semi-supervised anomaly detection methods across multiple sensor signal datasets.
\end{enumerate}

The structure of the remainder of the paper is as follows: Section 2 introduces relevant work and identifies existing challenges. Section 3 offers a formal definition of the problems tackled in this study. Section 4 provides a detailed description of the proposed model. Experimental results and their analysis are presented in Section 5.

\section{Related work}
We begin with an overview of anomaly detection, deep reinforcement learning, and causal inference.

\subsection{Anomaly Detection}
Anomaly detection algorithms can be categorized into three types, depending on the nature of data labeling: supervised, semi-supervised, and unsupervised anomaly detection methods. Supervised anomaly detection methods \cite{9965739-supervised,chen2022supervised, pang2021toward} rely heavily on the labels. However, the labels are difficult to obtain, so this type of method is unsuitable for most anomaly detection scenarios. Unsupervised anomaly detection methods \cite{XI2023110209-kbs,miao2024reconstruction-ipm,yoon2021self} do not require labeled data and can detect various anomalies. While these methods effectively prevent bias against known anomalies, they encounter challenges in identifying anomaly samples from normal samples because of the absence of prior knowledge about the true anomalies, especially when the anomaly patterns are very similar to the normal patterns.

Compared to unsupervised methods, Semi-supervised anomaly detection methods are designed for more practical scenarios where the dataset comprises numerous unlabeled training samples and a small number of labeled anomaly samples \cite{10102264-semisupervised,LIU2024111196-kbs}. These methods use partially labeled anomalies to improve the accuracy of detection. This study focuses on the task of semi-supervised anomaly detection in tabular form and considers the presence of novel unlabeled anomalies in the test set. Existing semi-supervised anomaly detection methods typically incorporate unsupervised and supervised learning techniques as a way to balance the limitations of each. One class of methods is based on distance/density approaches, which utilize known labeled data to establish baseline assumptions for distinguishing between normal and abnormal data and compute anomaly scores on unlabeled data \cite{pang2019deep}. Another class of methods first learns features from normal data and uses the learned features to supervise the model for distinguishing normal and abnormal samples \cite{LIU2024111196-kbs}. For example, one class of methods uses the Autoencoder or the generative adversarial networks (GAN) to determine the degree of abnormality of unknown points based on the reconstruction error \cite{TAN2024111533-kbs,zhang2023realistic,zhao2022ae}, and another class uses feature engineering means to fit specific abnormal features \cite{pang2023deep,xu2023rosas-ipm}. However, insufficient labeling information and limitations in the model assumptions (e.g., anomalies are rare or anomalies are significantly different in the feature space) make these methods still not efficiently utilize known anomalies to explore contaminated unlabeled data for potential unlabeled unknown anomalies effectively.

\subsection{Deep Reinforcement learning}
Deep reinforcement learning has demonstrated its power in many domains, such as game control, autonomous driving, and other more complex decision-making tasks \cite{adamczyk2023utilizing,huang2023efficient}. DRL is a learning approach in which a learning subject (called the "agent") learns how to complete a task or reach a goal by interacting with the environment. DRL's essence lies in optimizing decision-making by balancing exploration, the agent's capacity to refine the policy, and exploitation, the agent's use of acquired knowledge. This involves navigating unknown environments to maximize long-term rewards. Notably, DRL has been adeptly applied to anomaly detection by some researchers \cite{chen2023deep, pang2021toward，ijcai2023p19}. However, these algorithms still have a series of problems, such as the problem of the incentive mechanisms for incomplete feedback capacity(e.g., simply setting the decision reward to 0, 1, and -1), the problem of tendencies towards local optimality in \cite{pang2021toward}, and limitations with fixed threshold settings in \cite{chen2023deep}.

To address these issues, we proposed several measures. Illustrated in Figure \ref{fig1d}, the sampling strategy based on historical similarity aims to diversify the agent's exposure to different data points, enhancing exploration efficiency and reducing repetitive sampling. Moreover, to heighten the accuracy of DRL in anomaly detection, we replace fixed threshold methods with an adaptive threshold smoothing strategy. Lastly, to refine the DRL's decision-making accuracy, we propose a decision reward mechanism that makes it possible to adjust the value of the reward feedback according to the environment in which the agent is located. This strategy allows the agent to receive more comprehensive reward information rather than the incomplete reward information given in previous approaches. More in-depth analysis related to these strategies is presented in the ablation study section.
\subsection{Causal Inference}
As previously highlighted, a pivotal aspect of this study is eliminating confounding factors from samples to extract the intrinsic feature for which causal inference is ideally suited. Most existing deep learning models still focus on "learning from associated inference," which can lead to learning bias due to confounding factors in the sample \cite{adhikari2024discovering,ahuja2023interventional}. In contrast, causal inference focuses on "learning from causal inference," which seeks to remove confounding correlations from the sample by studying the effects of variables that change for some reason, avoiding irrelevant bias. Counterfactual intervention (CI) is often used to solve this problem. CI refers to using observations contrary to the original facts to replace the original facts in the observation process for causal discovery and further use of backdoor formulation \cite{pearl2013structural} to remove confounding correlations in the original data. Currently, CI has been widely used in various fields such as reinforcement learning \cite{gao2023fast} and computer vision \cite{lin2022causal}.

 Unlike previous work, we focus on causal inference between dimensional features in a sample rather than causality between historical observations, i.e., intra-point causality rather than inter-point causality. The reason for doing so is that this work considers tabular data and does not need to consider contextual relationships between data. To achieve this, we first embed the features of each dimension of the data points using the graph attention mechanism to obtain the graph embedding relationships between the features. Following this, our custom-built Causal Feature Extractor module, integrated with CI, facilitates the extraction of causal features from samples, and this process will be described in detail in the subsequent sections.
\section{Problem Formulation}

\subsection{Anomaly Decision-making Interactive Environment for Sensor Signals}

Ensuring the reliability of sensor signals in smart manufacturing is critical. Anomalies in sensor data can indicate potential faults or inefficiencies. Our objective is to train an agent using a training dataset $D$, which contains two initial training subdatasets: $D_a$, comprising labeled anomaly sensor signal data but smaller in size, and $D_u$, containing a larger volume of unlabeled sensor signal data. The agent's task is to assign an anomaly score to each sensor signal data point. The aim is to devise an anomaly scoring function, $\phi(\cdot) : \mathbb{R^{n}} \rightarrow \mathbb{R}$, ensuring that for any anomaly data point $o_i$ and normal data point $o_j$, the condition $\phi(o_i) > \phi(o_j)$ holds true, where $o_i,o_j \in D$.

Similar to conventional reinforcement learning, anomaly detection agents must interact with their environment to learn decision-making processes. To facilitate this, we develop the Anomaly Decision-making Interactive Environment (ADIE), which encompasses the following types of sample sets:

\begin{itemize}
    \item \textbf{Anomaly Data Pool $A$}: A data pool consisting of labeled anomaly data.
    \item \textbf{Unknown Data Pool $U$}: A data pool consisting of unlabeled data pending detection.
    \item \textbf{Temporary Data Pool $T$}: A buffer data pool consisting of temporary data obtained from the detection process.
\end{itemize}

$A$ and $U$ originate from the initial training datasets $D_a$ and $D_u$, respectively. At the start, $T$ contains no samples, but the sample counts in all three sets will fluctuate throughout the model iteration process. This dynamic is governed by two key parameters: the anomaly threshold (\emph{TH}) and the anomaly confidence threshold (\emph{TC}). \emph{TH}, a hyperparameter, varies with the environment and is initially set at 0.8. Each data point begins with a \emph{TC} of 0. If an observed point \textbf{\emph{o}} is provisionally deemed an anomaly by the agent, i.e., if $\phi(o_i) > \emph{TH}$, its \emph{TC} is incremented by 1, and the point is added to the temporary sample pool $T$. A point's \emph{TC} is also increased by 1 if it surpasses the sample pool's value. Once a point's \emph{TC} exceeds the set threshold, it is classified as an anomaly and transferred to the anomaly data pool $A$; otherwise, it remains in $T$. Additionally, if data in $T$ is assessed as normal, the point’s \emph{TC} is decreased by 1. Should the \emph{TC} of a point drop to 0, it is relocated to the sample set $U$. Data points not initially identified as abnormal by the agents will continue to stay in $U$.

\subsection{Anomaly Detection Agent}

The primary purpose of the ADIE is to select sensor signal data for analysis from the sample pool using a predetermined sampling strategy and present them as states to the agent. Upon the agent's decision-making and feedback to ADIE, ADIE must then provide appropriate reward information to the agent based on the reward strategy, facilitating updates to its scoring policy. To enhance the agent's decision-making comprehensiveness, we employ the Soft Actor-Critic (SAC) reinforcement learning algorithm, suited for continuous action spaces. SAC, an Actor-Critic-based algorithm, promotes exploration by incorporating entropy as an additional reward component, thereby encouraging the development of a robust anomaly scoring policy. In our study, the action space for the SAC agent ranges from $[0,1]$, with values closer to 1 indicating a higher likelihood of the tested sample being an anomaly. SAC's objective is to maximize the combination of cumulative returns and entropy. Entropy, reflecting the policy's randomness, spurs the algorithm to explore a broader range of behaviors. The objective function of SAC is represented by the equation:

\begin{equation}
    J(\pi_{\phi}) = \mathbb{E}_{(o_t, a_t) \sim \rho_{\pi_{\phi}}} \left[ \sum_{t} \gamma^t (R(o_t, a_t) + \alpha \mathcal{H}(\pi(\cdot|o_t))) \right]
\end{equation}

Where $\pi_{\phi}$ is the policy represented by a neural network that outputs a probability distribution of actions in a given state. $\rho_{\pi_{\phi}}$ is the distribution of the state and action under the policy $\pi_{\phi}$, $R(o_t, a_t)$ is the reward gained from taking action $a_t$ in the state $o_t$, $\gamma$ is the discount factor, $\mathcal{H}$ is the entropy, $\alpha$ is the entropy coefficient, which is used to control the importance of entropy in the objective function.

SAC uses a neural network to represent the policy $\pi_{\phi}(a|o)$, a policy network that outputs a probability distribution of actions in a given state. This network is trained to maximize the sum of Q-values and entropy. To minimize the learning bias of the model, SAC uses two Q-function networks and two state value function networks, including a $Q_{\theta_{1}}(o, a)$, a $Q_{\theta_{2}}(o, a)$, a target state value function $V'(o)$, and a state value function $V(o)$. The two Q-function networks estimate the cumulative payoff for the specific state and action. On the other hand, the state value function $V(o)$ represents the expected return that can be obtained by following the policy $\pi$ in the state $o$. At each training step, SAC updates the policy network, the two Q-function networks, the target value function network, and the state value function network. The policy network is updated to maximize the sum of the Q-value and entropy; the Q-function networks are updated to minimize the mean-square error between the Q-value and the target Q-value computed through Bellman's equation; and the state value function is updated to minimize the mean-square error between it and the expectation of the Q-value minus the entropy weights. At the end of the training, an optimal scoring policy will be obtained: $\pi^{*} = \arg\max_{(o_t, a_t)} \mathbb{E}$.

\subsection{Counterfactual Causal Inference}
Traditional reinforcement learning primarily relies on \textit{Correlation} rather than \textit{Causation}. However, relying only on correlation in complex and dynamic environments may lead to poor performance or incorrect decisions. In contrast, in \textit{Causal Reinforcement Learning}, the model understands the causal relationships between variables through causal inference, which provides insight into the true impact of a particular action on the environment. This approach sheds light on the genuine impact of actions on the environment, facilitating more effective and adaptive decision-making. Within causal inference, random variables in a causal graph \( G \) (a directed acyclic graph) are typically categorized into two groups: a set of covariates \( X \), represented as \( \{X_1, \ldots, X_n\} \), and an outcome variable \( Y \). To estimate causal effects, an intervention operation, denoted as \( \text{do}(X_j = x) \), is commonly utilized, setting the covariate \( X_j \) to a specific value \( x \). The causal effect of \( \text{do}(X_j = x) \) is then estimated using the backdoor formulation.

\begin{figure}[ht]
    \centering
    \subfloat[]{
        \includegraphics[width=0.3\textwidth]{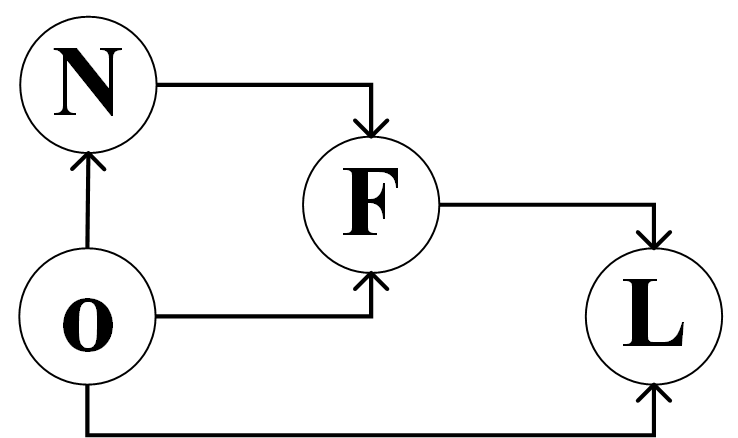}
        \label{fig2a}
    }
    \subfloat[]{
        \includegraphics[width=0.3\textwidth]{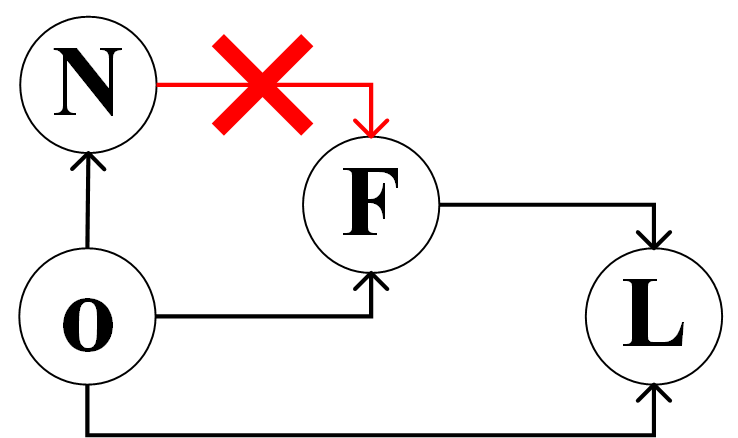}
        \label{fig2b}
    }
    \caption{The constructed causal graph. Figure(a) and Figure(b) represent the causal graph and the causal graph after using the counterfactual intervention.}
    \label{fig2}
\end{figure}
A central concept in causal inference is the backdoor path, which is crucial for identifying and addressing potential confounders among variables. Backdoor formulation operates within these paths, striving to negate the unwanted effects of confounders by obstructing backdoor paths. To illustrate this, we have constructed a directed acyclic graph(as shown in Figure \ref{fig2a}) that clearly depicts the causal transfer process. In this graph, \( O \) represents the original observational data, \( N \) symbolizes confounding noise features within it, \( F \) stands for the intrinsic causal feature, and \( L \) indicates the model's final decision outcome. The sequence \( O \rightarrow N \rightarrow F \rightarrow L \) forms a backdoor path in the graph. The confounding factor \( N = \{n_i\}^m_{i=1} \) in the original data then influences the model's decision outcome via this path, where \( m \) is the length of the confounder \( N \). 
\begin{equation}
    P(L|\text{do}(O), F) = \sum_{n} P(L | O, F, N=n) P(n)
    \label{eq6}
\end{equation}
We utilize the backdoor formulation to counteract this effect, as shown in Equation (\ref{eq6}). Based on Pearl's theory \cite{pearl2009causal}, the proof procedure of this formula is elaborated as follows:

\begin{align*}
P(L|\text{do}(O),F) &= \sum_{i=1}^{m} P(L | \text{do}(O), F,N=n_i) P(N=n_i|\text{do}(O)) \\
&= \sum_{i=1}^{m} P(L | O, F,N=n_i) P(N=n_i|O) \\
&= \sum_{i=1}^{m} P(L | O, F,N=n_i) P(n_i)
\end{align*}

Based on this setting, as shown in Figure \ref{fig2b}, we can conveniently use the CI to eliminate the adverse effects caused by confounding factors. We denote the backdoor formulation after intervention by CI as \( P(L|\text{do}(O), F) = \sum_{n'} P(L | O, F, N=n') P(n') \), where the confounding factor \( N \) is changed to other values such as 0, mean, and maximum as a result of the counterfactual intervention. To circumvent the challenges in precisely estimating high-dimensional probability distributions like \( P(L|O, F, N=n_i) \) and \( P(n_i) \), we employ a neural network to model the relationship between the original outcome and the counterfactual intervention outcome as a way of eliminating confounders in the original features and constructing intrinsic causal feature that has a critical impact on \( L \). Consequently, this module blocks the pathway from \( N \) to \( F \), preventing \( N \)'s adverse impact on \( L \).

\section{The proposed Model}
\subsection{Overall structure of the Tri-CRLAD for Sensor Signals}
\begin{figure}
    \centering
    \includegraphics{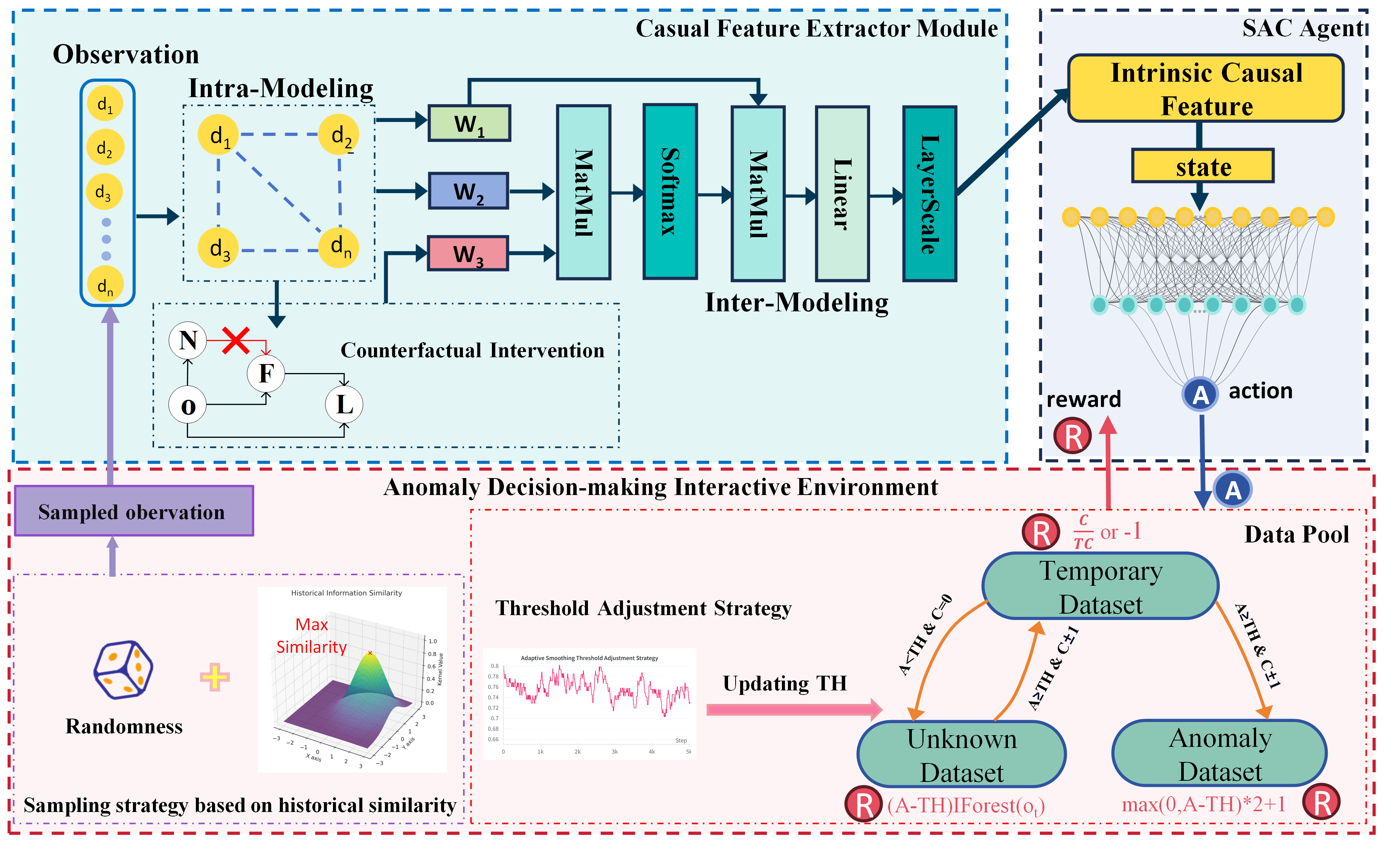}
    \caption{The overall structure of the Tri-CRLAD. To ensure the reliability of the collected sensor signal data, Tri-CRLAD comprises three principal components: the Causal Feature Extraction module, the DRL agent based on the SAC algorithm, and the ADIE module.}
    \label{fig3}
\end{figure}
To further illustrate how we ensure the reliability of the collected sensor signal data, this section provides a comprehensive description of the semi-supervised anomaly detection model we developed, named Tri-CRLAD. As depicted in Figure \ref{fig3}, the model's overall structure and data flow are illustrated. Tri-CRLAD comprises three principal components: the Causal Feature Extraction module, the DRL agent based on the SAC algorithm, and the ADIE module. The process unfolds as follows: the Causal Feature Extraction module extracts the intrinsic causal features from data points sourced from ADIE, and these features (\emph{States}) are then fed into the SAC. SAC, upon receiving these states, makes decisions (\emph{Actions}) and interacts with ADIE, which in turn provides feedback (\emph{Rewards}) based on the decisions to guide its scoring policy update. Next, we'll delve into the various interaction mechanisms designed specifically for ADIE.
\subsection{Triple Decision Support Mechanism}
To optimize the agent's adaptation to anomaly detection, we focus on the triple decision support mechanism within ADIE, which is based on historical information. This mechanism corresponds to the components outlined in the pink box in Figure \ref{fig3}. It includes a sampling strategy based on historical similarity, an adaptive threshold smoothing adjustment strategy, and an adaptive decision reward mechanism. In the following sections, we will thoroughly detail the individual roles of these three components and their collaborative synergies.
\subsubsection{Sampling Strategy Based on Historical Similarity}
As discussed in Section 3.1, one of ADIE's essential functions is supplying the agent with a data point for detection at each training step. Creating a sampling strategy encompassing a broad spectrum of scenarios within the dataset is vital. This breadth of sampling is necessary for the agent to gain a comprehensive understanding of the data and to effectively identify more anomalies within the unlabeled data, thereby significantly enhancing the agent's robustness. To achieve this, we have developed a sampling strategy based on historical similarity. This strategy is categorized into two types, each tailored to different data pool categories. Additionally, the roles of the various data pools are distinctly defined: the \( A \) data pool stores prior knowledge, the \( T \) data pool temporarily holds detected points for further analysis to increase detection accuracy, and the \( U \) data pool contains samples awaiting detection. During each training session, ADIE selects a data pool to sample from based on predetermined probabilities (set in this study at 0.3, 0.3, and 0.4 for the \( A \), \( T \), and \( U \) data pools). The selected data pool is denoted as \(D_s\). Accordingly, for different data pools, we implement the following two sampling methods:
\begin{enumerate}[1)]
    \item For data in \( A \) and \( T \), randomly sample data points from these two data pools as data points to be detected.
    \item For the data in \( U \), it is important to improve the efficacy of the sampling strategy therein. As shown in Figure \ref{fig1c}, the distance-based sampling strategy leads to frequent adoption of the same sample points. To fully balance the core process of exploration and exploitation in DRL, this paper proposes a sampling strategy as shown in Equation (\ref{eq3}), in which a composite score (\( \text{Score}(o) \)) is calculated to determine the sampled points. Here, the "Score" is a quantitative index of each point \( o_i \in O\), reflecting the probability of the $o_i$ being selected, where O represents all candidate points. The score level depends on a combination of two main factors: similarity to historical states and randomness. Specifically, a higher combined score means: \ding{172} High similarity to previously encountered valuable states, contributing to more efficient decision-making by effectively utilizing past experience. This part corresponds to the first half of Equation (\ref{eq2}), which utilizes a Gaussian kernel to find the average similarity between the set of points to be detected and all the historical data. \ding{173} The composite score also incorporates randomness, i.e., the second half of Equation (\ref{eq2}) is obtained by uniformly distributing the random samples \( r(o) \). Thus, even if a point is not the most similar to the historical state, it may receive a higher composite score due to the contribution of the randomness score. This promotes exploration, i.e., trying out new states that may not be identical to past experiences but still have potential. As shown in Figure \ref{fig1d}, for the same number of iteration steps, the sampling results based on the present strategy are more diverse, effectively reducing the risk of model overfitting.
\begin{equation}
   o^*=\left\{
\begin{array}{lcl}
\arg\max_{o_i \in O} Score(o_i)     &      & D_{s}=U\\
 random\_choose(D_s)     &      & D_s \in \{A,T\}
\end{array} \right.
\label{eq3}
\end{equation}

\end{enumerate}
\begin{equation}
    Score(o)=(1-\alpha )\frac{1}{n}\sum_{i=0}^{n}\exp\left(-\frac{\|o - h_i\|^2}{2\sigma^2}\right) + \alpha r(o)
    \label{eq2}
\end{equation}

where $\alpha$ represents bias, $h_i$ stands for the \(i\)th historical data, $n$ is the number of historical data in the staging, and $\sigma$ is the variance of the historical data.

In summary, the overall sampling strategy is shown in the Equation (\ref{eq3}). The strategy considers both the use of randomness to improve the exploration of new states and the use of historical information for decision-making. In addition, the Gaussian kernel-based similarity computation is non-parametric, meaning the strategy does not rely on a specific parametric model, making it more broadly applicable across different datasets and contexts.

\subsubsection{Adaptive Threshold Smoothing Adjustment Strategy}
Prior approaches set the anomaly determination threshold (\( \emph{TH} \)) to a fixed value, which limits the flexibility of the model when facing different task environments. Firstly, recall the pre-assumed scenario: there are a certain number of anomaly samples to be detected in the unlabeled sample set. This means that as the detection process proceeds, the number of abnormal samples in \( U \) decreases, and the agent's perception of the anomalies changes. This is because as the proportion of normal samples in \( U \) becomes larger, the agent will be more exposed to normal samples rather than abnormal samples. In view of this, there is an urgent need for an abnormality determination threshold adjustment method that can adaptively change with the detection process. To solve this problem, the adaptive threshold smoothing adjustment strategy is proposed here as a simple but effective measure. First, as shown in Equation (\ref{eq4}), we set an adaptive adjustment factor (\emph{Adjustment\textsubscript{factor}}) that is adjusted according to the results of anomaly decisions over a past period of time, and it will adjust the size of the factor according to the ratio of the past decisions judged to be anomalies(\(ratio_{current}\)) and the target ratio of anomalies in the data(\(ratio_{target}\)). Then, the threshold is adjusted using this factor within a limited range, i.e., \( \emph{TH} = \emph{Adjustment\textsubscript{factor}}*\emph{TH} \). In this process, limiting the adjustment range of the factor prevents the threshold from fluctuating drastically and ensures the stability of the system. 
\begin{equation}
    Adjustment_{factor} = 1.0 + ratio_{current} - ratio_{target}
    \label{eq4}
\end{equation}

Overall, this strategy makes the adjustment of thresholds flexible and smooth, allowing the agent to better adapt to the ever-changing anomaly detection environment.
\subsubsection{Adaptive Decision Reward Mechanism}
Reward settings in interactive environments are crucial for the efficiency and performance of the agent. As mentioned earlier, the complexity and variability of anomaly detection environments imply that effective reward feedback should be comprehensive and flexible. Existing approaches usually use fixed and one-sided reward mechanisms, typically using fixed values like 0, 1, and -1 as the feedback for the state, resulting in the agent only learning a narrow scope of reward information. To realize a more flexible and comprehensive reward mechanism, we propose the adaptive decision reward mechanism as shown in Equation (\ref{eq5}), which can be broken down into the following components according to the data pool in which the detected point is located: 
\begin{enumerate}[1)]
\item This section is for data points in the \textbf{pool \(A\)}. Link the reward to the current state of the environment, i.e., incorporate the \( \emph{TH} \) that reflects the current environment into the reward mechanism. The agent's reward is not merely determined by surpassing a fixed criterion. Instead, it is dynamically adjusted based on the relative variance from a fluctuating threshold, offering a more nuanced and responsive evaluation of the agent's behavior. Thus, this method offers more fine-grained rewards, responsive to minor environmental shifts, and encourages the agent to adapt accordingly.

\item This section is for data points in the \textbf{pool \(T\)}. Consider the cumulative impact of historical behavior and introduce historical information in rewards. Where \( C \) in Equation (\ref{eq5}) represents the confidence level that the current point is defined as an anomaly, \( \emph{TC} \) is the confidence threshold, and the point is identified as an anomaly when \( C \geq TC \). ADIE increments or decrements the reward according to the counts in the temporary data, i.e., \(\frac{C}{TC}\) or -1. Positive feedback is given when the detected point \(o_t\) is considered more likely to be an anomaly (action is upper than TH), and negative feedback is given when the opposite is true (action is lower than TH). In this way, the agent learns from single behavior and long-term behavioral trends, with higher rewards when behaviors are consistent with long-term goals, contributing to a more stable and effective scoring policy. 

\item This section is for data points in the \textbf{pool \(U\)}. The key is to adjust reward values dynamically, especially when facing unknown or uncertain states in the \( U \) data pool. We use the unsupervised method(IForest) to fine-tune rewards. This fine-tuning considers not only the current action or state but also the state's relevance and significance in the overall environment, enhancing the agent's understanding of the complex environment. In addition, introducing tiny negative rewards(-0.01) promotes the exploration of unfamiliar states, preventing the agent from overly fixating on familiar scenarios and encouraging broader environmental exploration. 
\end{enumerate}

Through these measures, our reward mechanism is more sophisticated and responsive to environmental dynamics, thereby enhancing the agent's learning efficiency and adaptability in complex settings.
\begin{equation}
    \text{Reward} = \begin{cases} 
    2\max(0, \text{action} - TH) + 1, & \text{if } o_t \in \text{A} \\
    \frac{C}{TC}, & \text{if } o_t \in \text{T and action} \geq TH \\
    -1, & \text{if } o_t \in \text{T and action} < TH \\
    (\text{action} - TH) \cdot \text{IForest}(o_t), & \text{if } o_t \in \text{U} \\
    -0.01, & \text{otherwise}
    \end{cases}
    \label{eq5}
\end{equation}
\subsection{Causal SAC}
To find the essential causal feature using CI and raw sensor signals, we need to model the relationship between them. We divide this modeling process into two categories: intra-point relationship modeling and inter-point relationship modeling. Intra-point relationship modeling refers to the use of the graph attention mechanism to embed the dimensions of data points into the attention graph, to discover the relationship between inner dimensions. Inter-point relationship modeling refers to the use of the use of CI to remove the confounding factors in the original data to extract the intrinsic causal feature. As shown in the causal graph (Figure \ref{fig2}), \( N \) is the confounding factor that needs to be removed. First, we use the backdoor formula as shown in Equation (\ref{eq6}) for this causal inference process. Because the counterfactual data \( c \) required for CI is unknown, to utilize the counterfactual data for intervention (\( \text{do}(X=c) \)), we replace each feature dimension in the original data with the mean of all dimensions as the counterfactual data. As described in Section 3.3, the counterfactual data intervene in the causal graph through the backdoor formula, which can make the path \( N \rightarrow F \) be blocked, avoiding the unfavorable influence of \( N \) on \( L \). Then, to obtain the intrinsic causal feature \( F \), we construct the Causal Feature Extractor (CFE) module, which models the feature information of the original data with that of the counterfactual intervention and explores the causal relationship between the two features, to infer the intrinsic causal feature \( F \) associated with each data point. The process is shown in the following equation:
\begin{equation}
F = Linear(softmax(\frac{(W_1x_{c})^{T}(W_2x)}{\sqrt{D}})W_3x)*scaler
    \label{eq7}
\end{equation}
where \( x \) represents the original data features, \( x_c \) represents the features obtained after counterfactual intervention, \( W_1 \), \( W_2 \), and \( W_3 \) represent the weights of different quantities, \( D \) is a constant scaling factor for feature normalization, and scaler is a layer used to promote convergence.

Subsequently, we explore how to effectively integrate the intrinsic causal feature \( F \) into a DRL driven by the SAC algorithm. Our objective function, defined as \( J(\pi) \), is constructed based on the following equation:
\begin{equation}
    J(\pi) = \mathbb{E}_{(F, a_t) \sim \rho_\pi} \left[ \sum_{t} \gamma^t (R(F, a_t) + \alpha \mathcal{H}(\pi(\cdot|F))) \right] 
\end{equation}

To make SAC work with the intrinsic causal feature extraction module, we adopt the idea of self-supervision. Specifically, we design a special asymmetric L2 loss function (Equation (\ref{eq8})). This loss function verifies the accuracy of the intrinsic anomaly feature by comparing the self-generated anomaly scoring information \( A_{\text{causal}} \) with the anomaly scoring information \( A_{\text{SAC}} \) generated by the SAC algorithm. When the loss function gives higher weight to positive errors, it makes the penalty for the anomaly scoring approach below the target value \( A_{\text{SAC}} \) greater, which helps to prevent generating scores that are so low as to underestimate the anomalies at a given point.
 \begin{equation}
     asymmetric\_L2\_loss(u) = \frac{1}{N} \sum_{i=1}^{N} \left( \tau \cdot \max(0, u_i)^2 + (1 - \tau) \cdot \max(0, -u_{i})^2 \right) 
        \label{eq8}
\end{equation}
 Where \( u \) is the set of prediction errors, \( u_i \) denotes a single prediction error, \( \tau \) denotes the weights of positive and negative errors in the loss function, \( \max(0, u_i) \) denotes positive error, and \( \max(0, -u_i) \) denotes negative error.
\begin{algorithm}
\caption{ Triple-Assisted Causal Reinforcement Learning Anomaly Detector}
\label{alg1}
\begin{algorithmic}[1]
\State Initialize policy network $\pi_\phi(a|o)$ and critic networks $Q_{\theta_1}(o, a)$, $Q_{\theta_2}(o, a)$,value network $V$,target value network $V'$
\State Initialize replay buffer $\mathcal{M}$ and Adam optimizer with learning rate $lr$
\State Initialize CFE with parameters $\eta$  
\For{each episode}
    \For {each training step}
        \State Sample data $o$ by Eq.(\ref{eq3})
        \State Get causal feature $F \leftarrow CFE(o)$
        \State Select action $a \sim \pi_\phi(\cdot|F)$
        \State Execute action $a$ in the ADIE
        \State Get reward $r$ according to Eq.(\ref{eq5})
        \State Update the $A$, $T$, and $U$ data pools in the ADIE
        \State Sample next point $o'$ by Eq.(\ref{eq3})
        \State Store transition tuple $(o, a, r, o')$ in $\mathcal{M}$
    \EndFor
    \If{Size of $\mathcal{M}$ $>$ $Warmup Size$}
        \State Sample a minibatch of transitions$\mathcal {B}$ $(o, a, r, o')$ from $\mathcal{M}$
        
        \For{each sampled transition}
            \State $F \leftarrow CFE(o)$,$F' \leftarrow CFE(o')$
            \State Compute target value $y = r + \gamma V'(F')$
           
            \State Update $Q_{\theta_{1,2}}$ by: $\nabla_\phi \frac{1}{|\mathcal{B}|} \sum_{F} \left(y - Q_{\theta_i}(F, a)\right)^2$
            \State Update $V$ by:$\nabla_\phi \frac{1}{|\mathcal{B}|} \sum_{F} \left(Q_{\theta_i}(F, \pi_\phi(F)) - V(F)\right)^2$
  
            \State Update $\pi_\phi$ by: $\nabla_\phi \frac{1}{|\mathcal{B}|} \sum_{F} \left( \min_{i=1,2} Q_{\theta_i}(F, \pi_\phi(F)) - \alpha \log \pi_\phi(\pi_\phi(F)|F) \right)$
            \State Calculate loss by Eq.(\ref{eq9})
            \State Update $\eta \leftarrow optimizer(\eta,loss,lr)$
        \EndFor
        
        \State Softly update target value network $V'$ 
    \EndIf
\EndFor
\end{algorithmic}
\end{algorithm}
 Moreover, we also ensure that the counterfactual data belongs to the same distribution as the original data by calculating the Maximum Mean Discrepancy (MMD) \cite{gretton2012kernel} between the original data feature $F_{original}$ and the causal feature $F_{causal}$, which satisfies the key prerequisite for causal inference. In this way, the overall loss function (Equation (\ref{eq9})) not only coordinates the interplay between reinforcement learning and causal inference but also builds a bridge that makes possible comprehensive modeling and inference from raw data to the intrinsic causal feature. 
   \begin{equation}
Loss = MMD(F_{original},F_{causal})+asymmetric\_L2\_loss(A_{SAC}-A_{causal})
        \label{eq9}
 \end{equation}
 
 Algorithm \ref{alg1} illustrates how each component interacts within the Tri-CRLAD. In summary, this model provides a powerful tool for understanding and analyzing the causality of data deeply, opening up new research and application possibilities.

\section{Experiments}
This section will present the sensor signals dataset used in this work, the evaluation metrics, and the results of Tri-CRLAD compared to 9 baseline methods.
\subsection{Datasets}
To comprehensively assess the generalizability of the Tri-CRLAD, we selected a diverse set of single-anomaly and multiple-anomaly datasets in the field of sensor signals, as shown in Table \ref{tab1}. These datasets cover various application areas, including medical monitoring, health diagnostics, behavior monitoring, and satellite monitoring, ensuring a broad and diverse scope for our research.

The single-anomaly datasets—annthyroid, cardio, satellite, and satimage—feature anomaly data percentages ranging from 1.2\% to 31.6\%. These datasets were specifically chosen to evaluate the model's effectiveness in handling single-anomaly scenarios with varying proportions of anomalies. On the other hand, the multiple-anomaly datasets include Multi-annthyroid for disease diagnosis, cardio for fetal cardiotocogram diagnosis, and har for human activity recognition. The proportion of single-category anomalies in these datasets ranges from 2.5\% to 15\%. These diverse datasets, encompassing multiple types of abnormalities, provide a robust evaluation of the model's performance in complex scenarios involving sensor signals. Utilizing these sensor data enables the diagnosis of thyroid disease and the monitoring of cardiovascular health.

\begin{table}[ht]
    \centering
    \caption{Description of Datasets}
    \scalebox{0.8}{%
        \begin{tabular}{ccccl}
            \toprule
            \multicolumn{2}{c}{Dataset} & \multicolumn{2}{c}{Anomaly Size} & \multicolumn{1}{c}{\multirow{2}*{Data Sources}} \\
            \cmidrule{1-4}
            Name  &   D  &  Class  & Size  &  \\
            \midrule
            annthyroid & 6   &abnormal& 534 (7.4\%)   & \url{http://odds.cs.stonybrook.edu/annthyroid-dataset/} \\
            cardio  &   21  &abnormal& 176 (9.6\%)  & \url{http://odds.cs.stonybrook.edu/cardiotocogrpahy-dataset/} \\
            satellite & 36 &abnormal& 2036 (31.6\%) & \url{http://odds.cs.stonybrook.edu/satellite-dataset/} \\
            satimage2 & 36 &abnormal& 71 (1.2\%) & \url{http://odds.cs.stonybrook.edu/satimage-2-dataset/} \\
            \midrule
            Multi\_cardio & 21 & \begin{tabular}{@{}c@{}}suspect \\ pathologic \end{tabular}&\begin{tabular}{@{}c@{}}295 (13.9\%)\\  176 (8.28\%)\end{tabular} & \url{https://archive.ics.uci.edu/ml/datasets/Cardiotocography} \\
            Multi\_har & 561 & \begin{tabular}{@{}c@{}}upstairs \\ downstairs \end{tabular} & \begin{tabular}{@{}c@{}}1544 (15\%) \\ 1406 (13.7\%)\end{tabular} & \url{https://www.openml.org/d/1478} \\
            Multi\_annthyroid & 21 & \begin{tabular}{@{}c@{}}hypothyroid\\ subnormal \end{tabular}&\begin{tabular}{@{}c@{}}93 (2.5\%) \\ 191 (5.1\%)\end{tabular} & \url{https://www.openml.org/d/40497} \\
            \bottomrule
        \end{tabular}
    }
    \label{tab1}
\end{table}
\subsection{Implementation Details}
In terms of experimental design, we divided these datasets into training and testing sets in the ratios of 0.8 and 0.2. In addition, to ensure the consistency of the data, we used the minimax scaler to preprocess these datasets. We initialize the Tri-CRLAD with Xavier initialization and employ Adam as the optimizer. By default, Tri-CRLAD undergoes training over ten episodes, each comprising 5,000 steps, followed by 5,000 warm-up steps and 10,000 warm-up sizes. In the Soft Actor-Critic (SAC) framework, the actor and critic are assigned a learning rate of 5e-4. The target network within SAC is updated at every 20-step interval. We retain the original SAC's default settings for all other optimization parameters. The TH is initialized as 0.8 and updated at every 10-step interval. 

The reported results are averaged across five distinct seed experiments. Then, the training set is processed according to a specific anomalies ratio and contamination ratio to ensure that the training set contains only a small amount of known anomalies $D_{a}$ and a certain amount of anomalies $D_{u}$ in the unlabeled data pool. Such processing helps to simulate the common real-world situation where known anomalies are scarce, and there are potential unlabeled anomalies. 
\subsection{Evaluation Metrics}
We use AUC-ROC as the evaluation metric for the experiments, a commonly used evaluation method in anomaly detection that can effectively measure the model's performance in distinguishing normal and anomaly data. Finally, we show the performance results of the Tri-CRLAD model by averaging five randomized seed experiments. The optimal and sub-optimal results are indicated by bolding and underlining the experimental results.
\subsection{Baselines and Comparisons}
We have chosen various methods, including supervised, unsupervised, and semi-supervised learning, as our baseline methods for this experiment. These methods include the popular XGBoost \cite{chen2016xgboost} and Isolation Forest \cite{liu2008isolation}, as well as two methods in the same field as our work - DADS \cite{chen2023deep} and DPLAN \cite{pang2021toward}. Additionally, we have selected several other SOTA methods for semi-supervised learning, such as DeepSAD \cite{ruff2019deep}, DevNet \cite{pang2019deep}, SSAD \cite{gornitz2013toward} and STOC \cite{yoon2021self} method. Notably, we cite the experimental results in DADS and the STOC+GOAD \cite{bergman2020classification} set up in that paper with the results of STOC+IForest as the performance of STOC in this experiment. The experimental results for DeepSAD correspond to the performance of an upgraded version of this model (Hypersphere Classification (HSC) \cite{ruff2020rethinking}).

In this experimental section, we will show the robustness of the proposed model step by step. First, we tested the model's ability to perform in a single anomaly dataset under an anomalies ratio of 0.1 and a contamination ratio of 0.1. This part of the experiment aims to test the fundamental ability that should be possessed as an anomaly detection model, i.e., the ability to detect anomaly sensor signal data points accurately. As demonstrated in Table \ref{tab2}, Tri-CRLAD excels at utilizing a minimal number of labeled anomalies to effectively identify unlabeled anomaly data points within the \( D_u \).

\begin{table}[!ht]
\centering
\caption{Performance of Different Methods on Single-class of Anomaly Datasets}
\renewcommand{\arraystretch}{1.5} 
\scalebox{0.9}{
\begin{tabular}{l c c c c c}
\toprule
& \textbf{annthyroid} & \textbf{cardio} & \textbf{satellite} & \textbf{satimage2} & \textbf{Average} \\ \hline
\textbf{Supervised \cite{chen2016xgboost}} & \textbf{0.984} & 0.912 & 0.838 & 0.969 & \underline{0.926} \\ 
\textbf{Unsupervised \cite{liu2008isolation}} & 0.804 & 0.920 & 0.769 & 0.994 & 0.872 \\ 
\textbf{DeepSAD \cite{ruff2019deep}} & 0.894 & 0.879 & \textbf{0.903} & 0.980 & 0.914 \\ 
\textbf{DevNet \cite{pang2019deep}} & 0.832 & 0.967 & 0.845 & 0.974 & 0.905 \\ 
\textbf{SSAD \cite{gornitz2013toward}} & 0.739 & 0.898 & 0.859 & 0.973 & 0.867 \\ 
\textbf{STOC+GOAD \cite{chen2023deep}} & 0.610 & 0.945 & 0.775 & 0.953 & 0.821 \\ 
\textbf{STOC+IForest \cite{chen2023deep}} & 0.833 & 0.919 & 0.786 & \textbf{0.994} & 0.883 \\ 
\textbf{DADS \cite{chen2023deep}} & 0.881 & 0.973 & 0.822 & 0.990 & 0.916 \\ 
\textbf{DPLAN \cite{pang2021toward}} & 0.824 & 0.769 & 0.852 & 0.813 & 0.815 \\ 
\textbf{Tri-CRLAD} & \underline{0.954±0.09} & \textbf{0.981±0.006} & \underline{0.873±0.062} & \underline{0.975±0.012} & \textbf{0.946} \\ 
\bottomrule
\label{tab2}
\end{tabular}
}
\end{table}

\begin{figure}[h]
	\centering
	\subfloat[]{
		\includegraphics[width=0.30\linewidth,height=0.25\linewidth]{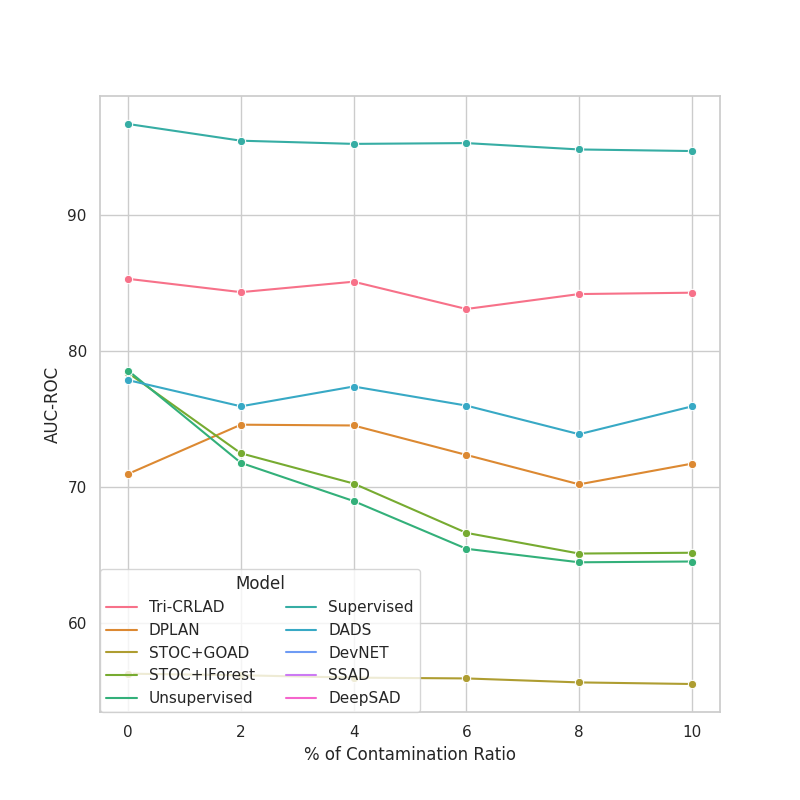}}
	\subfloat[]{
		\includegraphics[width=0.30\linewidth,height=0.25\linewidth]{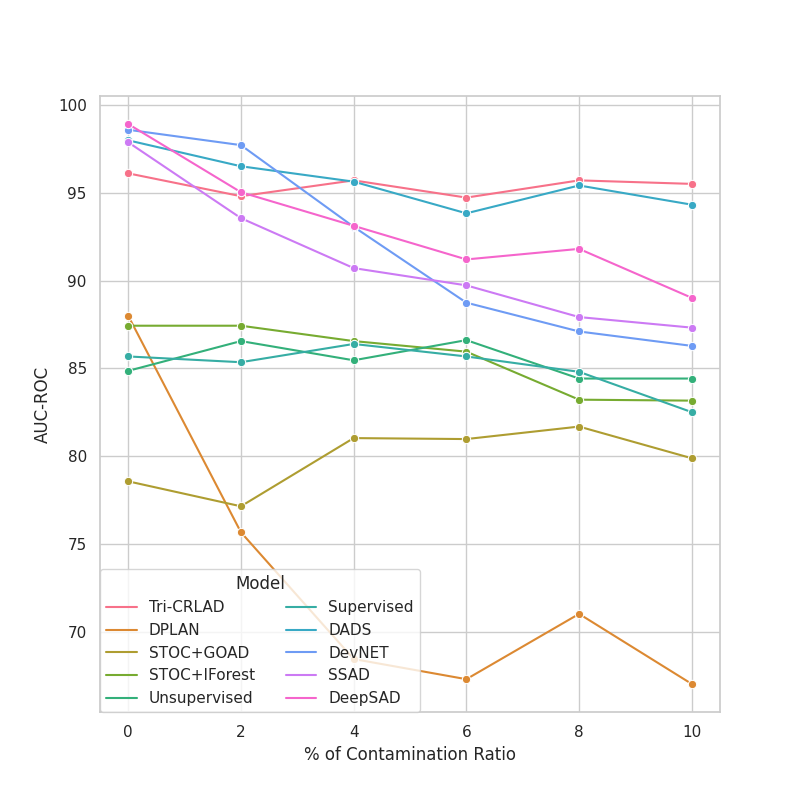}}
	\subfloat[]{
		\includegraphics[width=0.30\linewidth,height=0.25\linewidth]{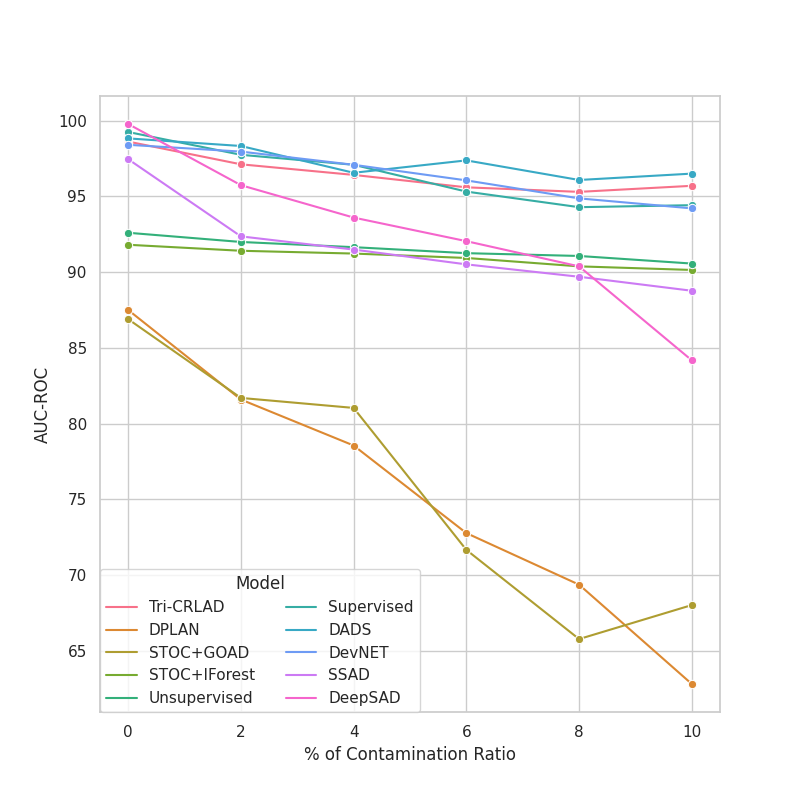}}\\
	\caption{Performance of Different Methods on Multi-class of anomaly Datasets in Scenario 1. Figure(a)(b)(c) shows the performance of Tri-CRLAD in Multi\_annthyroid, Multi\_cardio, and Multi\_har when the anomalies ratio is fixed and the contamination ratio is gradually increased}
    \label{fig5}
\end{figure}

Next, we further tested Tri-CRLAD's ability to perform in a multi-anomaly scenario. This scenario implies that an agent can detect anomalies containing a known class and other classes from unlabeled data when only a very small number of labeled anomalies within a class of anomalies are known, which is undoubtedly challenging. This part of the experiment can be divided into two parts, one in which the anomalies ratio is fixed and the contamination ratio is varied (Scenario 1), and the other in which the contamination ratio is fixed and the anomalies ratio is varied (Scenario 2). In Scenario 1, we fixed the anomaly ratio at 0.1 and adjusted the contamination ratio in increments of 0.02 within a range of [0, 0.1]. Figure \ref{fig5} shows the model's performance with different datasets, and Table \ref{tab3} exhibits the results when the anomalies ratio=0.1 and the contamination ratio=0.1. The outcomes of these experiments reveal that the model demonstrates considerable stability when facing gradual changes in contamination levels. This finding affirms the model's robustness and its capability to adapt effectively in environments characterized by multiple and challenging anomaly detection tasks within sensor signal data.

\begin{table}
\centering
\caption{Performance of Different Methods on Multi-class of Anomaly Datasets in Scenario 1}
\renewcommand{\arraystretch}{1.5} 
\scalebox{0.9}{
\begin{tabular}{l c c c c}
\toprule
& \textbf{Multi\_annthyroid} & \textbf{Multi\_cardio} & \textbf{Multi\_har} & \textbf{Average} \\ \hline
\textbf{Supervised \cite{chen2016xgboost}} & \textbf{0.949} & 0.826 & 0.944 & \underline{0.906} \\ 
\textbf{Unsupervised \cite{liu2008isolation}} & 0.645 & 0.844 & 0.906 & 0.798 \\ 
\textbf{DeepSAD \cite{ruff2019deep}} & 0.706 & 0.890 & 0.843 & 0.813 \\ 
\textbf{DevNet \cite{pang2019deep}} & 0.712 & 0.862 & 0.943 & 0.839 \\ 
\textbf{SSAD \cite{gornitz2013toward}} & 0.655 & 0.873 & 0.889 & 0.806 \\ 
\textbf{STOC+GOAD \cite{chen2023deep}} & 0.554 & 0.831 & 0.681 & 0.689 \\ 
\textbf{STOC+IForest \cite{chen2023deep}} & 0.651 & 0.799 & 0.903 & 0.784 \\ 
\textbf{DADS \cite{chen2023deep}} & 0.759 & \underline{0.943} & \textbf{0.965} & 0.889 \\ 
\textbf{DPLAN \cite{pang2021toward}} & 0.716 & 0.670 & 0.628 & 0.671 \\
\textbf{Tri-CRLAD} & \underline{0.842±0.013} & \textbf{0.955±0.007} & \underline{0.957±0.020} & \textbf{0.918} \\ 
\bottomrule
\label{tab3}
\end{tabular}
}
\end{table}

\begin{figure}[ht]
	\centering
	\subfloat[]{
		\includegraphics[width=0.30\linewidth,height=0.25\linewidth]{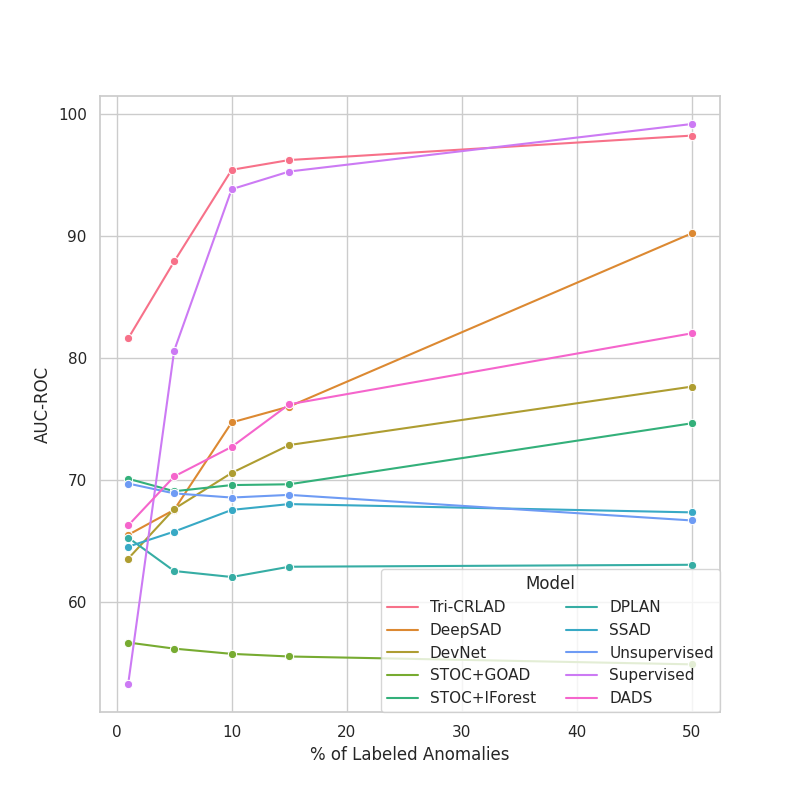}}
	\subfloat[]{
		\includegraphics[width=0.30\linewidth,height=0.25\linewidth]{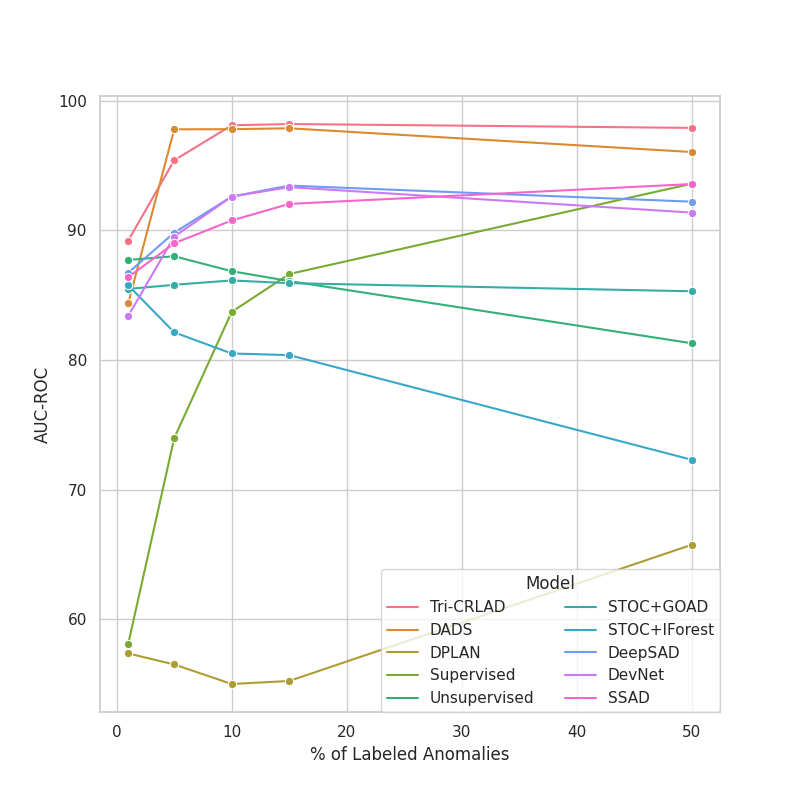}}
	\subfloat[]{
		\includegraphics[width=0.30\linewidth,height=0.25\linewidth]{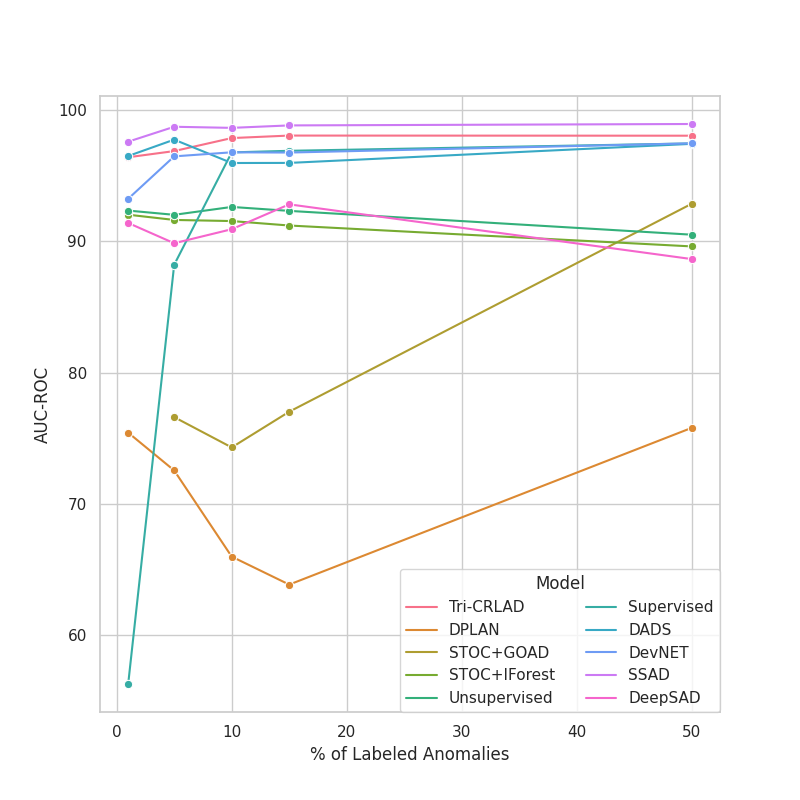}}\\
	\caption{Performance of Different Methods on Multi-class of anomaly Datasets in Scenario 2. Figure(a)(b)(c) shows the performance of Tri-CRLAD in Multi\_annthyroid, Multi\_cardio, and Multi\_har data when the contamination ratio is fixed and the anomalies ratio is gradually increased}
    \label{fig6}
\end{figure}

In Scenario 2, this part of the experiment fixes the contamination ratio at 0.04, and the anomalies ratio is varied in the range of 0.01, 0.05, 0.1, 0.15, and 0.5. The results presented in Table \ref{tab-Sce2} show the performance of the model for anomalies ratio=0.01, and Figure \ref{fig6} shows the performance of each model corresponding to each anomalies ratio. In the condition of anomalies ratio=0.01, the available prior knowledge is greatly reduced, and the capabilities of the various models show a significant decline. However, due to the effective collaboration between the CFE module and the triple decision support mechanism within ADIE, Tri-CRLAD exhibits resilience in this challenging scenario. Its maximum recession rate is limited to 15\% compared to the peak performance observed in this experimental segment. In contrast, other methods experienced a much steeper decline, with maximum recession rates reaching as high as 38\%. Consequently, considering the overall results, Tri-CRLAD maintains remarkable stability without significant fluctuations, further underscoring the model's exceptional robustness for dealing with sensor signals.
\begin{table}[!ht]
\centering
\caption{Performance of Different Methods on Multi-class of Anomaly Datasets in Scenario 2}
\renewcommand{\arraystretch}{1.5} 
\scalebox{0.9}{
\begin{tabular}{l c c c c}
\toprule
& \textbf{Multi\_annthyroid} & \textbf{Multi\_cardio} & \textbf{Multi\_har} & \textbf{Average} \\ \hline

\textbf{Supervised \cite{chen2016xgboost}} & 0.533 & 0.581 & 0.563 & 0.559 \\ 
\textbf{Unsupervised \cite{liu2008isolation}} & 0.697 & \ul{0.877} & 0.923 & 0.832 \\ 
\textbf{DeepSAD \cite{ruff2019deep}} & 0.646 & 0.867 & 0.914 & 0.809 \\
\textbf{DevNet \cite{pang2019deep}} & 0.636 & 0.834 & 0.932 & 0.801 \\ 
\textbf{SSAD \cite{gornitz2013toward}} & 0.646 & 0.864 & \textbf{0.976} & 0.829 \\ 
\textbf{STOC+GOAD \cite{chen2023deep}} & 0.567 & 0.855 & 0.638 & 0.687 \\ 
\textbf{STOC+IForest \cite{chen2023deep}} & 0.701 & 0.858 & 0.920 & 0.826 \\ 
\textbf{DADS \cite{chen2023deep}} & \ul{0.703} & 0.844 & \ul{0.965} & \ul{0.837} \\ 
\textbf{DPLAN \cite{pang2021toward}} & 0.653 & 0.574 & 0.755 & 0.661 \\ 
\textbf{Tri-CRLAD} & \textbf{0.816±0.035} & \textbf{0.892±0.004} & 0.964±0.002 & \textbf{0.891} \\ 
\bottomrule
\end{tabular}
}
\label{tab-Sce2}
\end{table}

The above experiments show that Tri-CRLAD shows a comprehensive lead in model stability and peak performance compared to similar work DPLAN and DADS. In addition, Tri-CRLAD performs more prominently in terms of robustness than supervised algorithms. Even in cases with very few known anomalies (e.g., an anomaly ratio of 0.01), the present model can still achieve good results without significant performance degradation, as in the case of supervised algorithms. 

Compared to other semi-supervised or unsupervised algorithms for non-reinforcement learning, Tri-CRLAD demonstrates a clear lead by its superior ability to utilize prior knowledge. Overall, Tri-CRLAD's success is attributed to its efficient sampling strategy, which motivates the agent to explore more extensively. Next, the intrinsic causal feature extracted through causal inference, combined with the adaptive decision reward mechanism, provides comprehensive learning information for the agent. Finally, with the help of the adaptive smoothing threshold adjustment strategy, the model can comprehensively perceive the learning situation to determine the abnormal sensor signal data accurately.
\subsection{Ablation Study}

This section will analyze the effects of several important components of the Tri-CRLAD, including the adaptive threshold smoothing strategy, the adaptive decision reward mechanism, and the CFE module. Since the superiority of our sampling strategy demonstrated in Figure \ref{fig1c} and \ref{fig1d} has already been explained in the previous section, we will not discuss this. To exemplify the superiority of these strategies constructed in this paper, we combine DPLAN with our ADIE, using this variant as a controller for this part of the experiment. As shown in Table \ref{tab 4}, we compared several variants of models with the Tri-CRLAD in three single-anomaly datasets set to an anomalies ratio of 0.1 and a contamination ratio of 0.1.

\begin{table}[!htb]
\centering
\caption{Impact of Components on Model Performance}
\renewcommand{\arraystretch}{1.5} 
\scalebox{0.9}{
\begin{tabular}{l c c c c }
\toprule
& \textbf{annthyroid} & \textbf{cardio} & \textbf{satellite} & \textbf{satimage2} \\ \hline

\textbf{DPLAN} & $0.826_{\pm0.035}$ & $0.768_{\pm0.008}$ & $0.852_{\pm0.011}$ & $0.825_{\pm0.022}$ \\ 
\textbf{DPLAN+TH} & $0.853_{\pm0.024}$ & $0.802_{\pm0.004}$ & $0.855_{\pm0.003}$ & $0.843_{\pm0.009}$ \\ 
\textbf{DPLAN+RW} & $0.881_{\pm0.006}$ & $0.876_{\pm0.017}$ & $0.853_{\pm0.008}$ & $0.855_{\pm0.006}$ \\ 
\textbf{DPLAN+causal} & $0.927_{\pm0.006}$ & $0.923_{\pm0.004}$ & $0.866_{\pm0.012}$ & $0.879_{\pm0.012}$ \\ \hline
\textbf{Maximum increase} & $\textbf{12.2\%}$ & \textbf{20.1\%} & \textbf{1.5\%} & \textbf{6.5\%} \\ 
\midrule
\textbf{Tri-CRLAD w/o TH} & $0.920_{\pm0.013}$ & $0.947_{\pm0.059}$ & $0.847_{\pm0.018}$ & $0.966_{\pm0.005}$ \\ 
\textbf{Tri-CRLAD w/o Rw} & $0.872_{\pm0.007}$ & $0.926_{\pm0.006}$ & $0.823_{\pm0.011}$ & $0.962_{\pm0.018}$ \\ 
\textbf{Tri-CRLAD w/o causal} & $0.841_{\pm0.13}$ & $0.937_{\pm0.006}$ & $0.818_{\pm0.102}$ & $0.958_{\pm0.018}$ \\ 
\textbf{Tri-CRLAD} & $0.954_{\pm0.09}$ & $0.981_{\pm0.006}$ & $0.873_{\pm0.062}$ & $0.975_{\pm0.012}$ \\
\bottomrule
\end{tabular}}
\label{tab 4}
\end{table}

\begin{figure}
    \centering

    \includegraphics[width=0.5\linewidth]{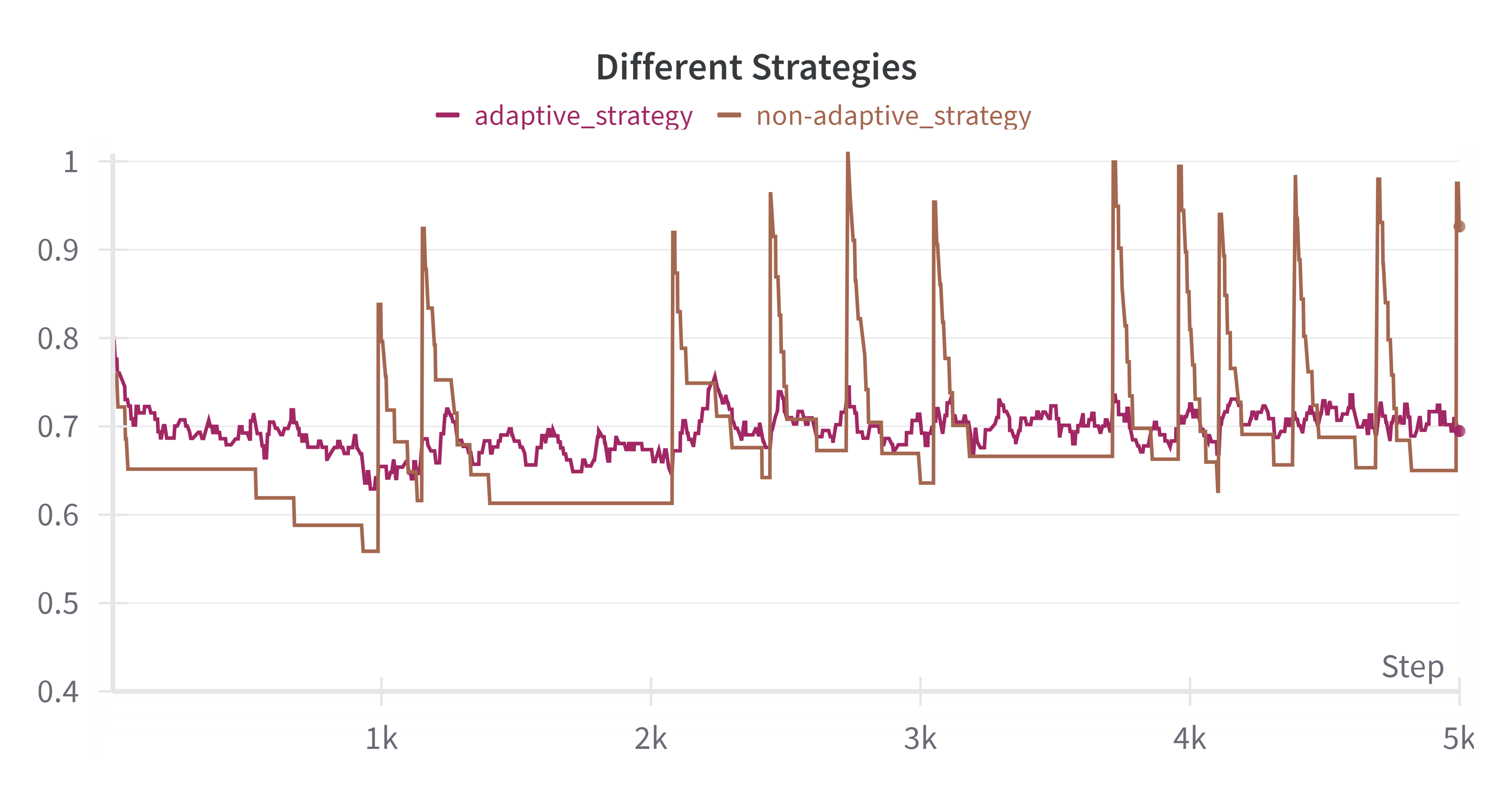}
    \caption{Performance of threshold changes when adjusting different strategies. The brown and red lines represent the threshold changes when using a non-adaptive adjustment strategy versus an adaptive smoothing threshold adjustment strategy.}

    \label{figTH}
\end{figure}

First, we analyze the role of the adaptive threshold smoothing strategy. The adaptive threshold smoothing strategy is designed to enable the agent to determine the abnormal data more accurately. As can be seen from the results in Table \ref{tab 4}, compared with the fixed thresholds, the adaptive changing thresholds enable the agent to detect anomalies more accurately. Meanwhile, to further demonstrate the superiority of the structure described in the method section, Figure \ref{figTH} demonstrates the difference in the changing trend under the non-adaptive adjustment strategy and the adaptive adjustment strategy constructed in this paper, and Table \ref{tab 4} also contains the impact of the two different results on the performance of Tri-CRLAD. This series of sufficient experiments shows that the adaptive threshold smoothing strategy constructed in this paper makes the adjustment of the threshold smoother on the one hand, making the model's decision more stable. On the other hand, the strategy improves the detection accuracy of the model, which can be better suited for various sensor signal tasks.

\begin{figure}[ht]
	\centering
	\subfloat[]{
		\includegraphics[width=0.5\linewidth]{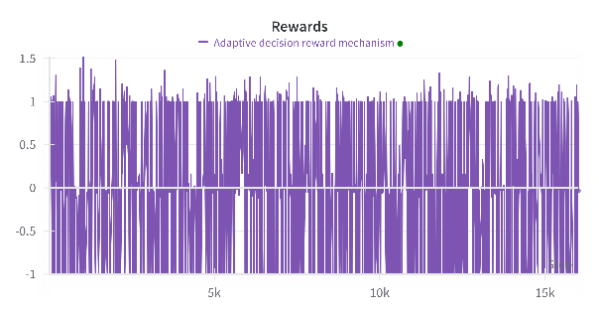}
        \label{fig7a}}
	\subfloat[]{
		\includegraphics[width=0.5\linewidth]{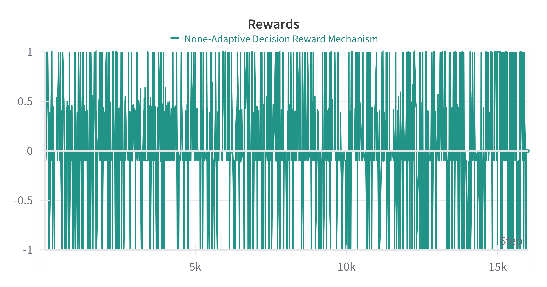}
        \label{fig7b}
        }
	\caption{Reward feedback when based on different reward mechanisms. Figure(a) and Figure(b) represent the feedback of the adaptive decision reward mechanism and the non-adaptive reward mechanism designed in the DADS}
    \label{fig7}
\end{figure}

Continuing the discussion is the adaptive decision reward mechanism, which is a key component in ADIE and greatly influences how well the agent learns. This strategy provides more comprehensive reward information to guide model learning. To demonstrate the advantages of this strategy, Figure \ref{fig7a} and \ref{fig7b} show the reward values generated by the adaptive and non-adaptive decision reward mechanisms. The model performance when using different reward mechanisms is provided in Table \ref{tab 4}. The non-adaptive reward mechanism is constructed in this case as described in DADS. This result shows that the reward mechanism provided by this strategy has two advantages over the non-adaptive decision reward mechanism: \ding{172} It can provide more comprehensive and fine-grained reward information. Figure \ref{fig7b} shows that the non-adaptive decision reward mechanism can provide a limited granularity of rewards with uneven positive and negative feedback.
In contrast, the adaptive decision reward mechanism can provide comprehensive positive and negative feedback information. \ding{173} Provide more intensive rewards. The rewards provided by the non-adaptive decision reward mechanism are relatively sparse, which is not conducive to the continuous learning of the agent based on the rewards. The present mechanism adjusts the value provided by the reward according to the relative difference with the dynamic threshold because it is denser, which enables the agent to adapt to the changes of the environment more flexibly and improves the learning efficiency of the model.

\begin{figure}[ht]
	\centering
	\subfloat[]{
		\includegraphics[width=0.3\linewidth]{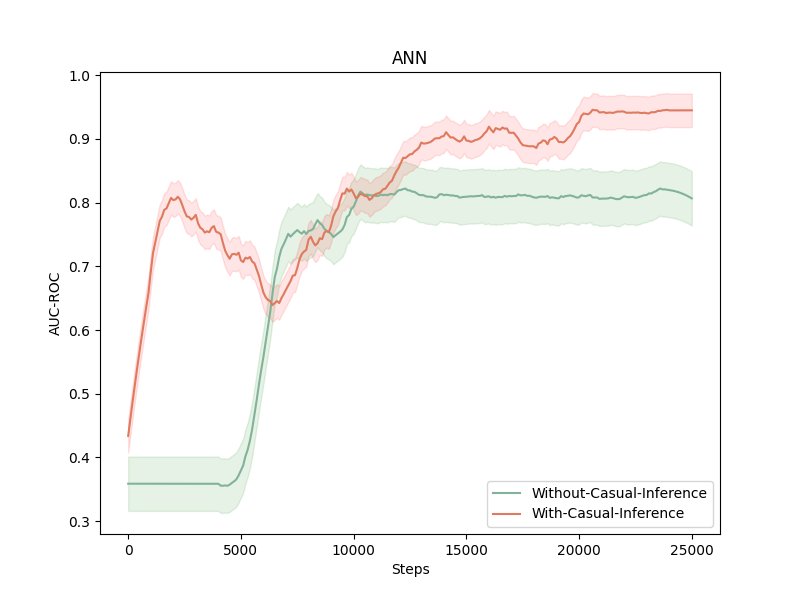}
        \label{fig8a}}
	\subfloat[]{
		\includegraphics[width=0.3\linewidth]{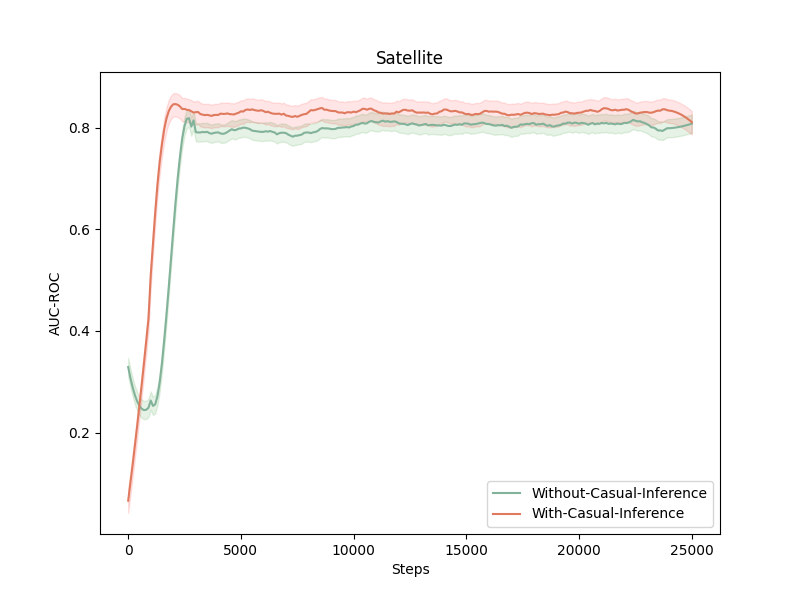}
        \label{fig8b}
        \includegraphics[width=0.3\linewidth]{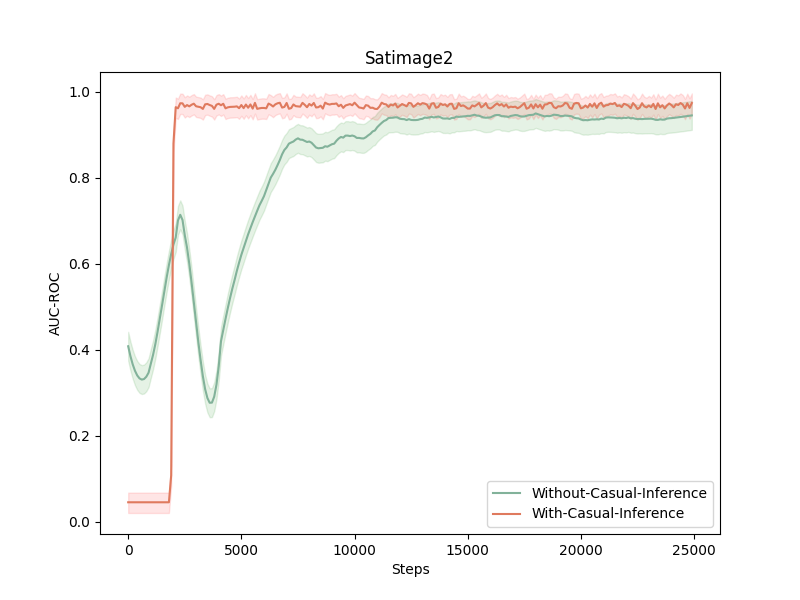}
        \label{fig8c}
        }
	\caption{Training curves for Tri-CRLAD compare scenarios where the causal inference module is utilized versus when it's substituted with the MLP. Where the green and red lines represent the training without the causal inference module and when the module is used.}
    \label{fig8}
\end{figure}

The last module discussed is the causal feature extractor module. This module utilizes counterfactual causal inference to extract the intrinsic causal feature in the data to improve the utilization of prior knowledge and reduce the learning bias of the agent due to confounding features in the data. It has already been mentioned in Figure \ref{fig1b} that the feature extracted according to this module enables the anomaly data points to be distinguished more clearly, demonstrating this module's effectiveness. In addition, the training curves of Tri-CRLAD with the causal inference module and with MLP instead of this module are shown in Figure \ref{fig8}, respectively. Combining the results in Table \ref{tab 4} with the training curves, it can be seen that the causal inference module enables the model to absorb prior knowledge more efficiently, thus achieving better learning results.

\section{Conclusion}
In the era of smart manufacturing, enhanced sensing and computing capabilities have made anomaly detection essential in various sensor-based applications. In this study, we propose the Tri-CRLAD model, a new approach to address the key challenges facing semi-supervised anomaly detection in the field of reinforcement learning. The model introduces causal reinforcement learning into semi-supervised anomaly detection and solves the problem of underutilizing prior knowledge by mining the intrinsic causal properties of data through counterfactual reasoning, which improves the ability of the agent to utilize prior knowledge. In addition, the model proposes a triple decision support mechanism based on historical information, which enhances the model's flexibility and improves the problem of insufficient reward feedback when interacting with the environment. Through these innovations, Tri-CRLAD can effectively respond to complex and changing sensor signal environments, demonstrating a wide range of potential applications. Ultimately, Tri-CRLAD demonstrates superior performance on multiple anomaly detection datasets in the sensor signal domain, meeting or exceeding existing state-of-the-art semi-supervised anomaly detection methods.

In future work, we will further improve the learning efficiency and stability of the reinforcement learning model in the anomaly detection process and investigate its application in the field of multivariate time series.

\section*{Acknowledgments}
This work was supported by the National Natural Science Foundation of China ( No.62172097), the Fujian Provincial Department of Science and Technology(No.2020H6011,2021Y0008). 

\section* {CRediT authorship contribution statement}	
\textbf{Xiangwei Chen}: Writing – original draft, Software, Methodology, Conceptualization, Data curation, Investigation. \textbf{Ruliang Xiao}: Methodology,Writing-review\& editing,Funding acquisition, Project administration,Supervision.\textbf{Zhixia Zeng}: Formal analysis, Investigation, Writing–review.  \textbf{Zhipeng Qiu}:Writing-review \& editing.  \textbf{Shi Zhang}: Writing-review. \textbf{Xin Du}: Writing-review.

\printcredits

\bibliographystyle{model1-num-names}
\bibliography{cas-refs}

\end{document}